\documentclass[a4paper,twocolumn]{article}

\usepackage[top=0.70in, bottom=0.50in, left=0.50in, right=0.50in]{geometry}

\usepackage{titlesec}
\usepackage{hyperref}
\usepackage{graphicx}
\usepackage{amsmath, amssymb, amsthm}

\renewenvironment{abstract}{\centerline{\bfseries Abstract}\begin{quote}\small}{\end{quote}}

\titleformat{\section}{\large\bfseries\raggedright}{\thesection}{1em}{}
\titleformat{\subsection}{\bfseries}{\thesubsection}{1em}{}

\title{\rule{\linewidth}{0.04in}\\
\bfseries\Large Combinatorial Modelling and Learning with Prediction Markets
\rule{\linewidth}{0.02in}}
\author{{\bfseries Jinli Hu}\\{\normalsize School of Informatics, The University of Edinburgh}}
\date{}

\begin{document}

\maketitle\thispagestyle{empty}

\begin{abstract}
Combining models in appropriate ways to achieve high performance is commonly seen in machine learning fields today. Although a large amount of combinatorial models have been created, little attention is drawn to the commons in different models and their connections. A general modelling technique is thus worth studying to understand model combination deeply and shed light on creating new models. Prediction markets show a promise of becoming such a generic, flexible combinatorial model. By reviewing on several popular combinatorial models and prediction market models, this paper aims to show how the market models can generalise different combinatorial stuctures and how they implement these popular combinatorial models in specific conditions. Besides, we will see among different market models, Storkey's \emph{Machine Learning Markets} provide more fundamental, generic modelling mechanisms than the others, and it has a significant appeal for both theoretical study and application.
\end{abstract}

{\small\tableofcontents}

\section{Introduction}

The models that are built from combining some individual models are popular in machine learning. In a combinatorial model, the individual models are components and a structure is given to combine the components appropriately, such as taking the average of all individuals or outputting the majority of the results. The structure is thus called the \emph{combinatorial structure}. The first impression of the combinatorial structures can be given by two popular models, \emph{ensemble learning}\cite{Dietterich2000} and \emph{graphical models}\cite{Bishop2006}.

\paragraph{Ensemble learning}
In machine learning, no individual model can always perform the best in all cases: one model is just suitable for a certain type of dataset or only part of the given dataset. However, we can always make good predictions by using multiple models instead of individual ones. This method is called ensemble learning. For example, for any dataset we compare and choose the algorithm that is the most suitable for this dataset, and then make predictions based on its result\cite{Littlestone1992, Montgomery2010}. By doing this our prediction will never be worse than the best algorithm on any dataset. Another example is the Netflix challenge\cite{Bell2007}. The basic idea shown of ensemble learning is treating different algorithms as components and combining them in an appropriate way.  An ensemble learning model can achieve high performance on the dataset without any prior knowledge of it, as long as its has an appropriate way of combining its components. In other words, with a good combinatorial structure an ensemble learning model can perform very well.

\paragraph{Graphical models}
In a graph the nodes or cliques can be treated as components and the edges represents how the components are combined, therefore it's natural to think a graph as a combinatorial model. A component in a graphical model always gives a marginal belief on few random variables, which on many occasions differs from the components in ensemble learning models that always give their beliefs on all variables.\\

Generalising the combinatorial structures has a long-term appeal for both theoretical study and application. It can help understand combinatorial models deeply and inspire us to build new models with novel structures.

In the latest years, prediction markets show a promise for becoming such a generic combinatorial model\cite{Arrow2008, Wolfers2004}. Prediction markets are born to be the information aggregators and they naturally combine the agents' beliefs through market mechanisms\cite{Chen2010}. The goods in the market are the contracts associated with certain outcomes of the future event\cite{Wolfers2006}. Each agent bets for the outcomes based on its own belief, by buying or selling some amount of the corresponding goods. When the market reaches equilibrium, the prices of goods are the aggregated beliefs for the outcomes associated with the goods.

A prediction market also has a combinatorial structure, and its agents are the components. Representing the structures in prediction markets have much flexibility, because there are quite many choices on the type of contracts, market mechanisms, and many ways of representing agent behaviours. Therefore different views on the market structures will result in different prediction market models, such as Storkey's \emph{Machine Learning Markets}, Lay's \emph{Artificial Prediction Markets}, Chen's \emph{Market Maker Prediction Markets}, etc.\\

This paper aims to show how the market mechanisms can generalise the modelling and implement these models in specific conditions. Besides, we will compare different market models. The criterion the paper uses to discuss modelling and compare different models is how fundamental and generic a model can be. Based on it, the paper thinks Storkey's \emph{Machine Learning Markets} model provides a better modelling mechanism than the others.

Our discussion is based on the \textbf{key papers} \cite{Geras2011, Storkey2011} and \cite{Barbu, Lay2009, Lay2010a}. The former two papers introduce Storkey's \emph{Machine Learning Markets} and the latter three Lay's \emph{Artificial Prediction Markets}. Both of them build a generic model successfully, but Machine Learning Markets model on more fundamental assumptions than the other. They also discuss the learning process briefly. Other papers such as \cite{Chen2010, Chen2007} and \cite{Hanson2007} introduce special agents, the market makers, and market scoring rules to represent the market mechanisms and to discuss the learning process. Although introducing market maker can help obtain some important results, it makes the prediction market less general than the former models, which do not introduce any special agents.

To show how generalisation is achieved, the paper will first review on few popular and typical models and the structures they hold before intruding prediction market models.\\

Chapter \ref{sec:models} introduces the popular combinatorial models and makes a summary at the end. Chapter \ref{sec:concepts} prepares for the following discussion by introducing some basic concepts in prediction markets. Chapter \ref{sec:modelling} discusses modelling and compares three market models. Chapter \ref{sec:learning} talks about learning briefly. Finally, Chapter \ref{sec:final} draws conclusion.

\section{Models that have combinatorial structures}
\label{sec:models}

Although only few models will be mentioned here, their structures are quite typical, such as weighted average of the beliefs, product of the beliefs.

\subsection{Boosting}
Boosting is a class of algorithms\cite{Schapire2003}. The idea is run the weak learner (who has many weak classifiers) on reweighed training data, then let learned classifiers vote. Boosting is an ensemble learning example. There are many implementations of boosting, some famous ones are \emph{AdaBoost}\cite{Freund1997} and \emph{Random Forrest}\cite{Breiman2001}.
\paragraph{AdaBoost}
The model of AdaBoost is simply weighted average,
\begin{equation}\label{eq:adaboost}
f(\mathbf{x}) = \sum_{i}\alpha_{i}h_{i}(\mathbf{x})
\end{equation}
Where $\alpha_{i}$ is the weight for the weak classifier $h_{i}$. Sometimes, $h_{i}$ is also called basis, hypothesis or ``feature'', These names just reflect the different angles of views on AdaBoost. For this algorithm, learning seems more important than modelling. If it wants to achieve good performances, it should have good choices on the weights. Here PAC (Probably Approximately Correct) learning theory\cite{Valiant1984} supports AdaBoost to choose weights appropriately. Despite the complicated learning process, the structure of AdaBoost is quite simple.
\paragraph{Random Forrest}
The main difference between Random Forrest and other boosting algorithms is it introduces stochastic properties to the model. Because of this Random Forrest is not even treated as a boosting method. We won't dwell on the terminology issue. It's more important to see the connections between them. The structure of Random Forrest is even simpler, where the weights are all units,
\begin{equation}\label{eq:randomforest}
f(\mathbf{x}) = \sum_{i}h_{i}(\mathbf{x})
\end{equation}
However, although its structure seems less flexible without weights, Random Forrest also achieves good performances. The reason is this algorithm puts more efforts on choosing bases. In fact, each basis $h_{i}(\mathbf{x})$ is a tree grown on the training data and controlled by a random vector $\Theta_{i}$, $h_{i}(\mathbf{x}) = h_{i}(\mathbf{x}, \Theta_{i})$. Similarly, Random Forrest has a simple structure but a complex learning process.

\subsection{Mixture model}
Suppose there is a set of experts. Each expert performs well only on part of the whole data domain. If a model can give the outputs based on the most suitable experts for the data, the model can always achieve high performances. This kind of model is called \emph{mixture model}. One example of mixture models is \emph{mixture of experts}\cite{Jacobs1991}.
\paragraph{mixture of experts}
To construct the mixture structure, this model assigns different weights to each expert according to the data points. If the experts are good to explain the data point, their weights will be larger than others. Therefore the structure is,
\begin{equation}
f(\mathbf{x}) = \sum_{i}w_{i}(\mathbf{x})h_{i}(\mathbf{x})
\end{equation}
Here the weights can be treated as the posterior knowledge given the observations. One special case in mixture of experts is mixture of Gaussians, where each expert is a Gaussian distribution and the corresponding weight $w_{i}(\mathbf{x})$ is its responsibility for the observed data. It's worth noting that one graphical model, the \emph{hidden space model}\cite{Bishop2006}, can represent this structure.

\subsection{Product of experts}
\begin{figure}
\centering
\includegraphics{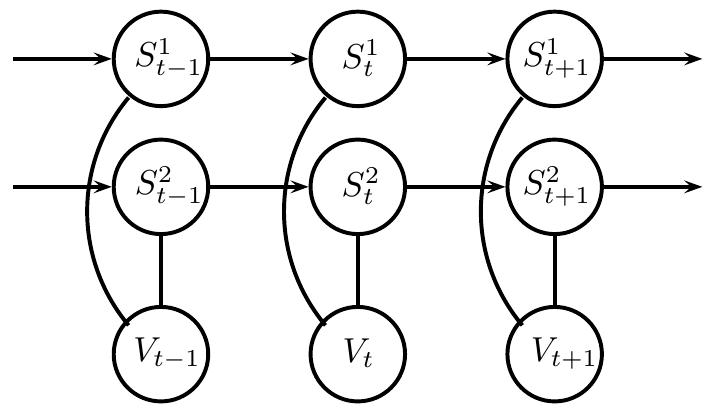}
\caption{The graphical model for Product of Hidden Markov Models}\label{fig:PoHMMs}
\end{figure}

This model has a combinatorial structure in a product form\cite{Hinton2002}. The product is always associated with graphs. The components are thus represented by nodes or cliques. The components give their own beliefs. When they are combined by different types of edges (directed or undirected), the final output are always in the form of the product of these beliefs. One example is product of HMMs\cite{Jacobs1991}.
\paragraph{product of HMMs}
The graphical model for product of HMMs is a mixture. It contains both directed edges and undirected edges (Figure \ref{fig:PoHMMs}).
The probabilities of variables in this case is complex. However, in graphs the joint probability distribution has a general \emph{product} form. Denote $\mathbf{x}$ as the vector which contains several random variables. We have,
\begin{equation}\label{eq:prob}
P(\mathbf{x}) = \frac{1}{Z}\prod_{C\in\mathcal{C}}\psi_{C}(x_{C})
\end{equation}
Here $\mathcal{C}$ is the set of cliques in the graph, $C\in\mathcal{C}$ is a certain clique and $x_{C}$ is the variables contained in the clique. The final output of the model are always derived from the joint distribution, by either doing marginalization or introducing evidence. For example, using (\ref{eq:prob}) we can write the probability of observed variable $V$ in this model\cite{Jacobs1991}, which we will omit here.

\subsection{Summary}
So far this paper has introduced several models and the combinatorial structures they represent. There are some common features in these structures. It's helpful to summarise these features before we discuss the generalisation of the structures.

In combinatorial structures, a component is an agent which has its own belief on the data. Different components (agents) act differently, by mapping the data to different probability distributions. To show this idea more clearly, we consider a \emph{sample space} $\Omega$ and its $\sigma$-field $\mathfrak{F}$, and suppose the data are drawn from this space $(\Omega, \mathfrak{F})$. What the agents actually do is they give their own probability measures on $(\Omega, \mathfrak{F})$. Thus agent $i$ has its own probability space $(\Omega, \mathfrak{F}, P_{i})$.

Combining these agents is actually a process of finding an appropriate map $\mathcal{F}: \{P_{i}\}\mapsto P$, where $P$ is still a probability measure on $(\Omega, \mathfrak{F})$. We've already seen some examples of $\mathcal{F}$. In boosting it is the weighted average. Now because of the much flexible definition, $\mathcal{F}$ should not be restricted to few forms such as weighted average or product, although only these structures have been seen in this paper so far.

Sometimes it's necessary to consider the situation that some agents may have their beliefs defined on the subspace of $(\Omega, \mathfrak{F})$. For example, in product models, the beliefs of agents are only defined on the cliques; they are also called \emph{local beliefs} or \emph{marginal beliefs}. There are two ways of interpreting these beliefs. The first one is introducing the subspace $(\Omega_{i}, \mathfrak{F}_{i})$. Now the probability space of agent $i$ is $(\Omega_{i}, \mathfrak{F}_{i}, P_{i})$, and thus $P_{i}(\Omega_{i}) = 1$. The second one is ``slicing'' the probability measure using \emph{Dirac} $\delta$, so that the probability of an event can be positive only if it is in $(\Omega_{i}, \mathfrak{F}_{i})$. However, compared with the first way, the subspace is not explicitly defined.

Suppose the data is drawn from the space $(\Omega_{i}, \mathfrak{F}_{i}, P^{T})$, where $P^{T}$ is the true probability distribution of the data, the learning process aims to give $P$ which is as close to $P^{T}$ as possible. In combinatorial structure, $P$ is constructed by the beliefs of agents $\{P_{i}\}$. Therefore we can think $\{P_{i}\}$ as bases and they form a hypothesis space. That's why in AdaBoost a component is also called basis or hypothesis. Besides, it helps understand modelling and learning in combinatorial structures.

\section{Basic Concepts in Prediction Markets}
\label{sec:concepts}

To fill the gap between the economics and machine learning, the paper will first introduce some basic concepts in Prediction Markets.

\subsection{Definitions}
\newtheorem{market}{Definition}
\begin{market}
A market is a mechanism for the exchange of goods. The market itself is neutral with respect to the goods or the trades. The market itself cannot acquire or owe goods. Unless explicitly stated otherwise, perfect liquidity is assumed and there is no transaction fee. All participants in the market use the same currency for the purposes of trade.
\end{market}

\begin{market}
The market equilibrium is a price point for which all agents acting in the market are satisfied with their trades and do not want to trade any further.
\end{market}

\begin{market}
Prediction market (also predictive market, information market) is a speculative market created for the purpose of making predictions. Current prices in market equilibrium can be interpreted as predictions of the probability of an event or the expected value of the parameter. One good is used as currency and each of the remaining goods is a bet on a paticular outcome of a future occurrence. A bet pays off if and only if the paticular outcome accoiciated with this bet actually occurs.
\end{market}

\paragraph{Types of Prediction Market}
We can design different types of bets (contracts) to realise different types of preditions in the prediction markets (Table \ref{tab:type}). For the purpose of predicting based on probabilities, we would like to choose the \emph{winner-take-all} type.
\begin{table*}
\centering
\begin{tabular}{lp{0.4\textwidth}p{0.2\textwidth}}
Contract & Details (with the example that Liberty Party win the vote) & Reveals market expectation of \ldots\\
\hline\\
Winner-take-all & Costs $\pounds p$. Pays $\pounds 1$ if and only if the Party wins $y\%$. Bid according to value of $\pounds p$. & Probability that event $y$ occurs, $p(y)$\\
\\
Index & Pays $\pounds 1$ for every percentage point won by the Party. & Mean value of outcome $y$: $E[y]$\\
\\
Spead & Costs $\pounds 1$. Pays $\pounds 2$ if the percentage $y > y^*$. Pays $\pounds 0$ otherwise. Bid according to the value of $y^*$ .& Median value of $y$\\
\end{tabular}
\caption{Different contract types in prediction markets}\label{tab:type}
\end{table*}

\begin{market}
Machine learning market (or artificial prediction market) is a special type of prediction market. Market participants are classifiers, which are also called agents. The market uses the winner-take-all contract. In the market the agents bet for the outcomes of the future events. They buy and sell bets based on their own beliefs. Prices in market equilibrium will estimate probabilities over the outcomes. 
\end{market}

\begin{market}
Utility is a measure of satisfaction, referring to the total satisfaction received by a consumer from consuming a good or service. A utility function is defined on the current wealth, and maps the satisfaction to a set of ordinal numbers, for which the common choice is $\mathbb{R}$. Any function is appropriate as long as it keeps the same ordinals.
\end{market}

\begin{market}
A buying function (also betting function) represents how much an agent is willing to pay for a good (or how many contracts it would like to buy) depending on its price.
\end{market}

\subsection{Notations}
For the consistency this paper follows the notations in \cite{Storkey2011} and \cite{Geras2011}.
\begin{itemize}
\item An outcome is an event defined on the sample space with a proper $\sigma$-field $(\Omega, \mathfrak{F})$. This space is mapped to a random vector $\mathbf{x}$ and each out come is denoted by $\mathbf{y}$. In prediction markets each outcome is a good. Goods are enumerated by $k = 1,2,\ldots,N_{G}$. In our discussion, $\{\mathbf{y}_{k}\}$ are mutually exclusive. For the sake of simplicity, $k$ can also denote the good $\{\mathbf{y}_{k}\}$. If there is only one random variable, the vector $\mathbf{x}$ will reduce to variable, and $N_{G}$ is the number of its outcomes. If there are $N$ variables in $\mathbf{x}$, the number of outcomes will increase exponentially on $N$.
\item The price of good $k$ is denoted by $c_{k}$. $\mathbf{c} = (c_{1}, c_{2}, \ldots, c_{N_{G}})^{T}$ denotes the price vector. $\sum_{k}c_{k} = 1$ maths its probability meaning.
\item The agents are enumerated by $i = 1, 2, \ldots, N_{A}$. The wealth of agent $i$ is denoted by $W_{i}$. The beliefs of agents on $\mathbf{x}$ are denoted by $P_{i}(\mathbf{x})$. So the belief of agent $i$ on good $k$ is $P_{i}(k)$, $\sum_{k}P_{i}(k) = 1$.
\item Stockholding of agent $i$ in good $k$ is denoted by $s_{ik}$. Negative stockholding indicates the agent sells the good, so $s_{ik}$ can be a negative value. $\mathbf{s}_{i} = (s_{i1}, s_{i2}, \ldots, s_{iN_{G}})^{T}$ denotes the vector of stockholding of agent $i$.
\item The utility of agent $i$ is denoted by $U_{i}$. The buying function of agent $i$ is denoted by $s_{ik}(W_{i},\mathbf{c})$.
\item If $\mathbf{x}$ contains multiple variables, they are enumerated by $j = 1, 2, \ldots, J$. As we discussed in Chapter \ref{sec:models}, we can introduce subspaces or cliques to make the beliefs defined only on only part of the variables in $\mathbf{x}$. The subspace of agent $i$ is denoted by $S_{i}$, and the variables in $S$ is denoted by $\mathbf{x}^{S_{i}}$, $\mathbf{x}^{S_{i}} = \{\mathbf{x}_{j}\mid\mathbf{x}_{j}\in S_{i}\}$. The agent's belief is denoted by $P_{i}(\mathbf{y}^{S_{i}})$. $\{\mathbf{y}^{S_{i}}\}$ are the outcomes of $\mathbf{x}^{S_{i}}$.
\item The training points are enumerated by $t = 1, 2, \ldots, T$. The $t$-th point is denoted by $D^{t}$.
\end{itemize}

\section{Modelling with Prediction Markets}
\label{sec:modelling}

The basic idea of modelling is that markets interpret the agents' behaviours in an appropriate way, and describing how they interact with each other. Then the connection between market prices and beliefs reveal the relationship between markets and other model structures.

\subsection{General market mechanism}
The prices of goods are determined by the \emph{equilibrium status of the market}. Different agents interact with each other in the way that they buy (sell) their preferred amount of goods to (from) others. Their behaviours are interpreted by the buying functions. When the market reaches the equilibrium, the supply matches the demand, thus no one would like to trade any more goods.
\begin{equation}\label{eq:equilibrium}
\sum_{i=1}^{N_{A}}\mathbf{s}_{i}(W_{i}, \mathbf{c}) = 0
\end{equation}
Note that $\mathbf{s}_{i}(W_{i}, \mathbf{c})$, so there are $N_{G}$ equations for $N_{G}$ goods. Substitute all buying functions into (\ref{eq:equilibrium}) and we can the solve the prices for $N_{G}$ goods. The prices are the aggregated probability distribution on the future events.

In many situations calculating the market equilibrium using (\ref{eq:equilibrium}) is difficult. For the simplicity of  numerical calculation, \cite{Geras2011} gives a score function which is called \emph{market equilibrium function}.
\begin{equation}\label{eq:score}
E(\mathbf{c}) = \sum_{k}\left(\sum_{i}s_{ik}(W_{i}, \mathbf{c})\right)^{2}
\end{equation}
Here $E(\mathbf{c}) \geq 0$, and the equality holds if and only if (\ref{eq:equilibrium}) holds. Therefore minimising $E(\mathbf{c})$ will give the market equilibrium.

Therefore, market equilibrium is the general market mechanism that aggregates the individual beliefs in the market. $\mathbf{s}_{i}(W_{i}, \mathbf{c})$ has not been specified here. Implement the buying function in different ways will give different models.

\subsection{Model buying function -- Artificial Prediction Markets}
One way to implement the buying function is we model the buying function directly\cite{Barbu, Lay2009, Lay2010a}. These papers call buying function ``betting function'', and they think we can choose its form arbitrarily. These papers define the buying function in a factorial form,
\begin{equation}
s_{ik}(W_{i}, \mathbf{c}) = W_{i}\phi_{i}(k, \mathbf{c})
\end{equation}
Where $\phi_{i}(k, \mathbf{c})$ means the proportion of wealth the agent would like to use. These papers give three types of the proportion functions.
\begin{itemize}
\item Constant proportion functions
\begin{equation}
\phi_{i}(k, \mathbf{c}) = P_{i}(k)
\end{equation}
\item Linear proportion functions
\begin{equation}
\phi_{i}(k, \mathbf{c}) = (1 - c_{k})P_{i}(k)
\end{equation}
\item Aggressive proportion functions
\begin{equation}
\phi_{i}(k, \mathbf{c}) =
\begin{cases}
1 & \text{if $c_{k} \leq P_{i}(k) - \epsilon$}\\
0 &\text{if $c_{k} \geq P_{i}(k)$}\\
\frac{P_{i}(k) - c_{k}}{\epsilon} & \text{otherwise}
\end{cases}
\end{equation}
\end{itemize}
The advantage of modelling the buying functions directly is, we can suppose some functions with simple forms and can obtain the results analytically. For example, use constant proportion function and (\ref{eq:equilibrium}), we have
\begin{equation}\label{eq:apm}
c_{k} = \frac{\sum_{i}W_{i}P_{i}(k)}{\sum_{i}W_{i}}, \quad k = 1,2,\ldots,N_{G}
\end{equation}
This is exactly the weighted average of all beliefs, and it interprets the structure of AdaBoost in (\ref{eq:adaboost}). Especially, if the weights are all the same, it interprets Random Forest in (\ref{eq:randomforest}).

\subsection{Model utility -- Machine Learning Markets}
The drawback of this formulation is, however, the buy function doesn't always have a factorial form. Besides, from the economics point of view, the buying function is \emph{not the foundation} to interpret the agent's behaviour\cite{Varian2010}. Instead, utility function is the one. According to \cite{Varian2010}, buying function is derived from the agent's rational behaviour, that the agent always wants to maximise its utility subject to its budget constraint. Therefore the utility function is defined on the wealth. In prediction markets, the utility has uncertainty. The expected utility function (which is called \emph{Von Neumann-Morgenstern utility}) can be written as\cite{Storkey2011},
\begin{equation}\label{eq:expected}
\begin{aligned}
\mathbb{E}[U_{i}] &= \sum_{k=1}^{N_{G}}P_{i}(k)U_{i}(W_{i} - \mathbf{s}_{i}^{T}\mathbf{c} + s_{ik})\\
&\qquad i = 1,2,\ldots,N_{A}
\end{aligned}
\end{equation}
This formulation introduces a free degree: agents can change their stockholdings by making risk free trades, namely changing the holding from $\mathbf{s}_{i}$ to $\mathbf{s}_{i} + \alpha\mathbf{1}$, and these trades don't affect their utilities (these utilities keep the same ordinals). We can introduce the \emph{gauge} or \emph{standardisation constraint} to eliminate this free degree. One choice of the gauge is,
\begin{equation}\label{eq:gauge}
\mathbf{s}_{i}^{T}\mathbf{c} = 0
\end{equation}
Then (\ref{eq:expected}) is rewritten as,
\begin{equation}
\begin{aligned}
\mathbb{E}[U_{i}] &= \sum_{k=1}^{N_{G}}P_{i}(k)U_{i}(W_{i} + s_{ik})\\
&\quad s.t.\quad \mathbf{s}_{i}^{T}\mathbf{c}=0, \quad i = 1,2,\ldots,N_{A}
\end{aligned}
\end{equation}
It's worth noting that, (\ref{eq:expected}) doesn't guarantee the invariances under translation, because
\begin{equation}
\begin{aligned}
U_{i}(W_{i} - (\mathbf{s}_{i}+&\alpha\mathbf{1})^{T}\mathbf{c} + s_{ik}) =\\
&U_{i}(W_{i} - \alpha - \mathbf{s}_{i}^{T}\mathbf{c} + s_{ik})
\end{aligned}
\end{equation}
The invariances can hold only when $U_{i}(x) = U_{i}(x+t)$, which may not always be met. In \cite{Chen2010} and \cite{Chen2007} the author constructs a cost function that can always hold the translational invariances.

Maximising the utility function gives the buying function. For agent $i$, the buying function is,
\begin{equation}
\mathbf{s}_{i}(W_{i}, \mathbf{c}) = \arg\max_{{s}_{i}} \mathbb{E}[U_{i}] \quad s.t.\quad \mathbf{s}_{i}^{T}\mathbf{c}=0
\end{equation}
Taking derivatives w.r.t each $s_{ik}$ to get the maximum. Use Lagrange multiplier to include the gauge and we have,
\begin{equation}
P_{i}(k)U_{i}'(W_{i} + s_{ik}) - \lambda_{i}c_{k} = 0
\end{equation}
Solve the $s_{ik} = s_{ik}(W_{i}, \mathbf{c}, \lambda_{i})$ from the above and combine it with the gauge (\ref{eq:gauge}), we can finally solve the buying function $s_{ik}(W_{i}, \mathbf{c})$. In \cite{Geras2011} the author gives three types of utility functions and derives the corresponding buying functions.
\begin{itemize}
\item Logarithmic
\begin{equation}\label{eq:log}
U_{log}(x) =
\begin{cases}
\log(x) & \text{if $x > 0$}\\
-\infty & \text{otherwise}
\end{cases}
\end{equation}
\begin{equation*}
s_{ik}(W_{i}, \mathbf{c}) = \frac{W_{i}(P_{i}(k) - c_{k})}{c_{k}}
\end{equation*}
This buying function has a linear form.
\item Exponential 
\begin{equation}\label{eq:exp}
U_{exp}(x) = -\exp(-x)
\end{equation}
\begin{equation*}
s_{ik}(W_{i}, \mathbf{c}) = \log P_{i}(k) - \log c_{k}
\end{equation*}
Because in exponential utility function we have $-e^{-W-x} = -e^{-W}e^{-x}$, so when taking derivatives the wealth term in the utility function is eliminated. Therefore the buying function do not depend on the wealth.
\item Isoelastic ($\eta > 0$)
\begin{equation}\label{eq:iso}
U_{iso}(x) = \frac{x^{1-\eta} - 1}{1 - \eta}
\end{equation}
\begin{equation*}
s_{ik}(W_{i}, \mathbf{c}) = W_{i}\left[\frac{\left(\frac{P_{i}(k)}{c_{k}}\right)^{1/\eta}}{\sum_{j}c_{j}\left(\frac{P_{i}(j)}{c_{j}}\right)^{1/\eta}} - 1\right]
\end{equation*}
Note that when $\eta \to 1$, it becomes the logarithmic case.
\end{itemize}

\subsection{Implementing combinatorial structures}
In former section, we have seen how prediction markets generally work to aggregate information and interpret it in prices of goods. Aggregation is determined by market equilibrium, which is based agents' behaviours described by their utilities. There is no restriction on choosing agents' utilities, so all agents can either have the same utility function, or their unique ones. The market whose agents share the same utilities is called \emph{homogeneous market}, otherwise it's called \emph{inhomogeneous market}. An inhomogeneous market is more general since it can become a homogeneous one by assigning the same utility to all agents. However, homogeneous market can implement these popular combinatorial structures we mentioned before and can even bring completely new structures, let alone the inhomogeneous market that may give much more outcomes. The prediction market models, Storkey's \emph{Machine Learning Markets}\cite{Geras2011, Storkey2011} and Lay's \emph{Artificial Prediction Markets}\cite{Barbu, Lay2009, Lay2010a}, both use homogeneous markets to form the combinatorial structures.

\paragraph{Homogeneous market with logarithmic utilities}
Using (\ref{eq:log}) and market equilibrium condition (\ref{eq:equilibrium}), we have,
\begin{equation}\label{eq:training}
c_{k} = \frac{\sum_{i}W_{i}P_{i}(k)}{\sum_{i}W_{i}}
\end{equation}
It's the same with (\ref{eq:apm}) which indicates that Machine Learning Markets can give the same results that Artificial Prediction Markets can give.

\paragraph{Homogeneous market with exponential utilities}
Using (\ref{eq:exp}) and market equilibrium condition (\ref{eq:equilibrium}), we have,
\begin{equation}
c_{k} \propto \prod_{i=1}^{N_{A}}P_{i}(k)^{1/N_{A}}
\end{equation}
It interprets the structure of Product of HMMs in (\ref{eq:prob}). Here every clique $C$ on which the beliefs are defined is a markov chain (Figure \ref{fig:PoHMMs}).

\paragraph{Homogeneous market with isoelastic utilities}
Using (\ref{eq:iso}) and market equilibrium condition (\ref{eq:equilibrium}), we have,
\begin{equation}\label{eq:isoprice}
c_{k} \propto \sum_{i=1}^{N_{A}}W_{i}\left[\frac{\left(\frac{P_{i}(k)}{c_{k}}\right)^{1/\eta}}{\sum_{j=1}^{N_{G}}c_{j}\left(\frac{P_{i}(j)}{c_{j}}\right)^{1/\eta}}\right]
\end{equation}
This is not a closed form because the right side contains prices. Besides, it's a \emph{novel} combinatorial structure, and all the models we ever discussed don't have such kind of structure. Despite the lack of model examples, we can infer some properties of this structure according to (\ref{eq:isoprice}).
\begin{itemize}
\item Similar to logarithmic case, the agents that have large weights (or that are more wealthy) $W_{i}$ contribute more to the market pricing. They would like to make the market price close to their beliefs. In economics, these agents are described to act like the ``price makers''\cite{Chen2010, Chen2007, Hanson2007}.
\item The personal belief is not so important as in logarithmic case. In stead, it is the \emph{relative belief} $P_{i}(k)/c_{k}$ that really affects the prices. If an agent is wealthy but his belief has a large deviation from others, the market will still treat it as an ``outlier'' and reduce its contribution.
\end{itemize}

\subsection{Agents with beliefs on subspaces}
We have discussed the agents that always have their beliefs on the whole random vector $\mathbf{x}$. This situation is true if $\mathbf{x}$ is actually a random variable, which means $\mathbf{x}$ contains only one entry. However, if there are multiple entries in $\mathbf{x}$, it will be more general to see the agents have beliefs only on part of $\mathbf{x}$. Introducting the agents with marginal beliefs make a model more general since the agents are not required to have their knowledge on all random variables now. As we mentioned in Chapter \ref{sec:models}, there are two ways to define the belief on the subspace.

The paper\cite{Storkey2011} introduces the subspace explicitly and defines the belief on it. Suppose $\mathbf{x}$ contains $J$ variables, $\mathbf{x} = (\mathbf{x}_{1}, \mathbf{x}_{2}, \ldots, \mathbf{x}_{J})^{T}$, and agent $i$ has its belief only on those variables in its subspace $S_{i}$. Then the belief is written as $P_{i}(\mathbf{y}^{S_{i}})$. $\{\mathbf{y}^{S_{i}}\}$ are the outcomes of the subspace variables $\mathbf{x}^{S_{i}}$. Because $P_{i}(\mathbf{y}^{S_{i}})$ is a marginal probability distribution, \cite{Storkey2011} calls the agents who have their beliefs on subspaces the \emph{marginal agents}.

The expected utility function for marginal agents is written as,
\begin{equation}
\begin{aligned}
\mathbb{E}[U_{i}] &= \sum_{\mathbf{y^{S_{i}}}}P_{i}(\mathbf{y}^{S_{i}})\times\\
&U_{i}(W_{i} - \sum_{\mathbf{y}^{S_{i}}}s_{i}(\mathbf{y}^{S_{i}})c(\mathbf{y}^{S_{i}}) + s_{i}(\mathbf{y}^{S_{i}}))
\end{aligned}
\end{equation}
Where $c(\mathbf{y}^{S_{i}}) = \sum_{\mathbf{y'}\backslash\mathbf{y'}^{S_{i}} = \mathbf{y}^{S_{i}}} c(\mathbf{y})$. If $S_{i}$ is exactly the whole space, then the above equation is back to (\ref{eq:expected}). There is no simple representation for the prices with marginal agents, but we know the market will give prices based on these marginal beliefs.

\subsection{Prediction markets and the map $\mathcal{F}:\{P_{i}\}\mapsto P$}
Flexible choices of utilities (buying functions) and beliefs make market models general. Again we can understand market models using the map. What the combinatorial structure provides is a map that make the agents' beliefs $\{P_{i}\}$ map to the aggregated belief $P$. The prediction market can implement a large number of combinatorial structures by choosing different utilities or buying functions. In former sections we have seen how the market results in those popular structures, such as homogeneous market with logarithmic/exponential utilities, and the structure that has not ever been used in any combinatorial models, such as homogeneous market with isoelastic utilities. The flexibility the prediction market has when it implements the map $\mathcal{F}:\{P_{i}\}\mapsto P$ shows that the prediction market is likely to be a generic combinatorial model.

One interesting question is how many maps on earth can be interpreted by the market. So far no work has been done, but solving this question will help refine this theory and thus it's worth further study.

\subsection{Another market mechanism -- Market Maker Prediction Markets}
Besides Storkey's \emph{Machine Learning Markets} and Lay's \emph{Artificial Prediction Markets}, another work in this field is done by Chen et al.\cite{Chen2010, Chen2007}. They introduce a market maker to help analyse good pricing and the bounded loss of learning. In \cite{Chen2007}, the authors prove the equivalence of three different market marker mechanisms: \emph{market scoring rule (MSR) market maker}, \emph{utility-based market maker} and \emph{cost function based market maker}. It's worth noting that, the utility-based market maker is pretty similar to the machine learning markets, where the market maker's behaviour is based on its expected utility function. However, instead of maximising the utility function, the agents in the markets aim to maximise their expected wealth subject to the invariance of the market maker's utility. Suppose there is only two agents, one is a trader (denoted by $t$)and another is the market maker (denoted by $m$). Then the market mechanism is written as,
\begin{equation}
\max_{\vec{m}} \sum_{k}P_{t}(k)(-m_{k}) \quad s.t. \quad \sum_{k}P_{m}(k)U_{m}(m_{k}) = C
\end{equation}
Where $m_{k}$ is the money the market maker spend on the good $k$. The money the trader spend is $-m_{k}$ for the assumption that the total wealth of two agents is $0$. $C$ is a constant.

Introducing a special agent has its own drawback: it makes the market \emph{less general} because in many cases there should be no any special agents in a market. Compared with Market Maker Prediction Market, the Machine Learning Markets model doesn't introduce any special agents and thus is better to be a generic model.

\section{Learning Process}
\label{sec:learning}

The problems of learning discussed here not only refer to the training, but also refer to the evaluation. Given one model, people are most curious about how good it can actually perform. For prediction markets both problems have not been completely solved yet. However, current results do show prediction market models have good performances, at lease from the Bayesian view.

\subsection{Training market models}
In \cite{Geras2011} the author discusses two ways of training the homogeneous market with logarithmic utilities, whose structure is represented by (\ref{eq:training}). Because agents' beliefs don't change, only their wealths $\{W_{i}\}$ keep updating during the training. Therefore the training is also called wealth updating in this paper. The two ways of wealth updating are \emph{online update} and \emph{batch update}. Online update means $\{W_{i}\}$ update after every training sample. Batch update means that each agent’s wealth is divided into equal pieces, one piece for each training point, and the updated wealth is the sum of all the updated pieces. Before training, we need choose an appropriate initial wealths. For example, we can choose uniform wealths $\{1/N_{A}\}$. Denote them as $\{W_{i}^{0}\}$. 
\paragraph{online update}
\begin{equation}
\begin{aligned}
W_{i}^{t+1} &= W_{i}^{t} - \mathbf{s}_{i}^{T}\mathbf{c} + s_{ik^{t}} = W_{i}^{t} + s_{ik^{t}}\\
&= W_{i}^{t} + \frac{W_{i}^{t}(P_{i}(k^{t}) - c_{k^{t}})}{c_{k^{t}}}\\
&= \frac{W_{i}^{t}P_{i}(k^{t})}{c_{k^{t}}}
\end{aligned}
\end{equation}
Where $k^{t}$ means the true outcome of training point $D^{t}$, and because of the standardisation constraint, $\mathbf{s}_{i}^{T}\mathbf{c} = 0$, we have $W_{i}^{t} - \mathbf{s}_{i}^{T}\mathbf{c} + s_{ik^{t}} = W_{i}^{t} + s_{ik^{t}}$. Train the market with $T$ data points, we will get the final wealth for each agent $W_{i} = W_{i}^{T}$.

\paragraph{batch update}
\begin{equation}
\begin{aligned}
W_{i} &= \sum_{t=1}^{T}\left[\frac{W_{i}^{0}}{T} + \frac{W_{i}^{0}}{T}\frac{(P_{i}(k^{t}) - c_{k^{t}})}{c_{k^{t}}}\right]\\
&= \frac{W_{i}^{0}}{T}\sum_{t=1}^{T}\frac{P_{i}(k^{t})}{c_{k^{t}}}
\end{aligned}
\end{equation}

\subsection{Bayesian view on training}
Now we think the agent's personal belief in another way. Suppose agents are represented by a random variables $\mathbf{z}$, which has $N_{A}$ outcomes $i = 1, 2, \dots, N_{A}$. The joint portability distribution over the random vector $(\mathbf{x}^{T}, \mathbf{z})^{T}$ is denoted by $P'(\mathbf{x}, \mathbf{z})$. Then each agent's belief $P_{i}(k)$ is the conditional probability $P'(\mathbf{x}=k|\mathbf{z}=i)$, namely we have $P_{i}(k) = P'(\mathbf{x}=k|\mathbf{z}=i)$. For simplicity we write $P'(k|i) = P'(\mathbf{x}=k|\mathbf{z}=i)$. In fact, we treat agents as the outcomes of a \emph{latent variable} $\mathbf{z}$. If we marginalise it on $\mathbf{z}$, we will obtain $P'(k)$.

The market prices form another probability distribution on $\mathbf{x}$, $c_{k} = P(k)$. Note that $P'$ and $P$ are two different probability measures on the sample space. $P'$ is directly obtained by the sum rule, while $P$ represents an arbitrary aggregation process. Only in one special case we have $P'=P$: the aggregation process is the (weighted) sum over all agents, which is the same as marginalization. The homogeneous market with logarithmic utilities just provide such a process. When $P'=P$ we have,
\begin{equation}
\begin{aligned}
P(k) &= P'(k) = \sum_{i}P'(k,i)\\
&= \sum_{i}P'(i)P'(k|i) = \sum_{i}P'(i)P_{i}(k)
\end{aligned}
\end{equation}
Where $P'_{i}$ can be treated as the weight of agent $i$, $W_{i} = P'_{i}$ (suppose $W_{i}$ has been normalised). This is exactly the \emph{Bayesian learning} in \cite{Bishop2006}. Now we consider the learning process and introduce training data.
\begin{equation}
P(k|D) = \sum_{i}P'(i|D)P_{i}(k|D)
\end{equation}
We see that $W_{i} = P'(i|D)$. So $W_{i}$ is the responsibility of agent $i$ for data $D$, or the posterior probability of agent $i$ given data. In Bayesian learning, as the number of data points increases, the posterior distribution $P'(\mathbf{z}|D)$ becomes sharp. After enough training points, the agent whose belief is the most close to the true probability distribution will gain the largest weight, while all the others have nearly zero weights. Therefore, in the homogeneous market with logarithmic utilities, finally the prices will be made by the best agent.

In fact we can understand this Bayesian process as a \emph{Bayesian model averaging}\cite{Montgomery2010}. It is often used to select the best algorithm for a given dataset, among a number of algorithms, but it can select only one algorithm\cite{Minka2000}. Strict deduction shows that, with infinite number of training points $T\to\infty$, if the hypothesis space contains the true probability distribution, Bayesian model averaging can select it out; if the hypothesis space doesn't contain the true one, then it will select the distribution that is the most close to true one.

Therefore, with enough points, the homogeneous market with logarithmic utilities will perform the same good as the best agent in the market. They have the same loss, which means how well the market can perform will depend on how good the best agent is.

\subsection{Further discussion}
Bayesian learning seems have solved the learning problem elegantly, but in fact it's too ideal. The so-called ``with enough points'' is a condition that can never be met, because we need $T\to\infty$ training points. It will be much more significant to answer how good performance the market can achieve with \emph{finite} or \emph{few} training points. In fact, this question is where some learning theories originate, such as \emph{Vapnik-Chervonenkis theory}.

For some models, such as boosting, this question has been well solved. So if the market mechanism results in boosting algorithms, we can know the performance of the market. However, we cannot know its performance if the market gives other combinatorial structures. What we want to know is whether it's possible to have a general idea on the performance of market, regardless of the combinatorial structure it represents.

Some work has done by Chen et al. and their discussion is based on introducing the market maker. In \cite{Chen2010} they show that ``\emph{any} cost function based prediction market with bounded loss can be interpreted as a no-regret learning algorithm''. Besides, in \cite{Chen2007} they have shown the equivalence between cost function based prediction market and utility based prediction market under some conditions. As we mentioned before, it's possible to connect the market marker model and Machine Learning Markets model. For example, we can specify an agent in the machine learning markets as a market maker. If it's true, the results of the market maker model can be used on Machine Learning Markets, making this model better and more general.

\section{Conclusion}
\label{sec:final}

Prediction markets have shown a huge potential of becoming a generic combinatorial models. In this review, we see how prediction markets provide flexible mechanisms to implement different combinatorial structures. By assigning specific utilities or buying functions to each agent and using market equilibrium condition, \emph{Artificial Prediction Markets} and \emph{Machine Learning Markets} can give not only the structures of some well-known models (logarithmic utilities implement averaging structure, and exponential utilities implement product structure), but also ones that have never been used (isoelastic utilities). More generally, prediction markets provide many optional choices of the map $\mathcal{F}:\{P_{i}\}\mapsto P$ where each $\mathcal{F}$ stands for a combinatorial structure. Therefore the prediction market models are general.

This paper has first compared two prediction market models, \emph{Artificial Prediction Markets}, which models the buying functions, and \emph{Machine Learning Markets}, which models the utilities. Both of them perform well in generalising combinatorial models. However, because utility is more conceptually fundamental than buying function and it can give buying function naturally, Machine Learning Markets model is more general and more elegant, than Artificial Prediction Market. Besides, the paper also discusses another market model, called Market Maker Prediction Markets. This model introduce a special agent, the market maker, to help analyse the market mechanism and pricing (namely learning) process. Admittedly some significant results are obtained. However, introducing the special agent makes it less generic than the former two models, which is a big problem. Therefore, based on the criterion that a better model should be more fundamental and more general, the paper holds the conclusion that Machine Learning Markets is the better one. A possible work might be trying to find the connections between Machine Learning Markets and Market Maker Prediction Markets, in order to use these significant results to refine Machine Learning Markets model.

A brief discussion on the learning process from Bayesian view shows that market models (Machine Learning Markets) do have high performances and thus are attractive to practitioners.

Apart from the modelling and learning theories, Machine Learning Markets model shows a promise in practical implementations. In the market all agents have the same priority, and their behaviours are independent with each other, only based on their own utilities. Parallelisation is therefore able to be used in the model, which shows a long-term appeal when people need to deal with more and more data.\\

In sum, prediction market models are becoming the generic combinatorial models. Among different market models, Machine Learning Markets has a particular promise of becoming the best market model, for it builds the model more general and elegantly, and gives attractive applications in the future.

\bibliographystyle{plain}
\bibliography{irrlibrary.bib}

\end{document}